\RequirePackage{iftex}
\RequirePDFTeX

\documentclass{article}

\usepackage[final]{corl_2026} % Camera-ready version.

\usepackage{url}
\usepackage{amsmath}
\usepackage{amssymb}
\usepackage{multirow}
\usepackage{multicol}
\usepackage{mathrsfs}
\usepackage{etoolbox}
\usepackage{xspace}
\usepackage{graphicx}
\usepackage{caption}
\usepackage{subcaption}
\usepackage{adjustbox}
\usepackage{xfrac}
\usepackage{listings}
\usepackage{float}
\usepackage{booktabs}
\usepackage{bm}
\usepackage{amsfonts}
\usepackage{enumitem}
\usepackage{bbding}
\usepackage{wrapfig}

\newcommand{\ours}[3]{RoboPocket}

\newcommand{\ie}{\textit{i}.\textit{e}., }

\usepackage{hyperref}
\usepackage{cleveref}
\crefname{figure}{Fig.}{Figs.}
\crefname{section}{Sec.}{Secs.}
\crefname{table}{Tab.}{Tabs.}
\crefname{algorithm}{Alg.}{Algs.}

\begin{document}

\title{RoboPocket: Improve Robot Policies Instantly\\ with Your Phone}

\author{
\bfseries Junjie Fang$^{1,3,*}$, Wendi Chen$^{1,2,*}$, Han Xue$^{1,3,*,\dagger}$, Fangyuan Zhou$^{1,2,*}$\\
\bfseries Tian Le$^{1,3}$, Yi Wang$^{1,2}$, Yuting Zhang$^{3}$, Jun Lv$^{3}$, Chuan Wen$^{1,\ddagger}$, Cewu Lu$^{1,2,3,\ddagger}$\\
\normalfont $^1$Shanghai Jiao Tong University, $^2$Shanghai Innovation Institute, $^3$Noematrix Ltd.\\
\normalfont $^*$Equal contribution, $^\dagger$Project lead, $^\ddagger$Corresponding authors\\
\normalfont\href{https://robo-pocket.github.io}{\textbf{robo-pocket.github.io}}
}

\maketitle

\begin{abstract}
Scaling robot learning is bottlenecked not only by data volume, but also by whether the data targets the policy's failures. Existing robot-free handheld interfaces enable portable in-the-wild data collection, yet remain open-loop: users record demonstrations without feedback from the current policy. Interactive methods such as DAgger provide such feedback by collecting corrective on-policy data, but require deploying imperfect policies on physical robots, making them costly, unsafe, and difficult to scale. We present \textbf{RoboPocket}, a smartphone-based system for \textit{robot-free policy iteration}. By combining on-device sensing, display, and communication with cloud-side policy inference, RoboPocket provides real-time collection guidance and visualizes the policy's predicted trajectory in augmented reality as \textit{AR Visual Foresight}. This allows users to identify policy weaknesses and collect targeted corrective data without deploying a physical robot. An asynchronous finetuning pipeline continuously incorporates new corrections and updates the policy within minutes, realizing a DAgger-inspired robot-free correction loop.
Across diverse real-world manipulation tasks, we validate RoboPocket as a scalable data engine and show that online corrective data collection improves sample efficiency by up to 2$\times$ over scaling offline demonstrations. Project page and videos: \href{https://robo-pocket.github.io}{\textbf{robo-pocket.github.io}}.
\end{abstract}


\section{Introduction}
\label{sec:intro}

Scaling has transformed computer vision and natural language processing, where massive data enables increasingly general models~\cite{kaplan2020scaling, hoffmann2022training}. In robotics, however, data remains a central bottleneck~\cite{o2024open}. Recent robot-free handheld interfaces~\cite{chi2024universal, generalist2025gen0, rdt2} mitigate this by decoupling data collection from hardware, enabling in-the-wild demonstrations. Yet, scaling imitation learning requires more than blindly collecting trajectories; the data must effectively cover states where the current policy fails.

This exposes a fundamental limitation: existing robot-free collection pipelines are mostly open-loop. Users record demonstrations without knowing if the data is valid, expands useful state coverage, or addresses the policy's weaknesses. Conversely, interactive methods like DAgger~\cite{ross2011reduction} directly target policy-induced failure states through corrective data, but require deploying imperfect policies on physical robots. This makes iteration costly, unsafe, and difficult to scale beyond controlled labs. Consequently, robot learning faces a critical trade-off: robot-free collection is scalable but policy-agnostic, while interactive correction is policy-aware but strictly robot-dependent.

\begin{figure}[t]
\centering
\includegraphics[width=\linewidth]{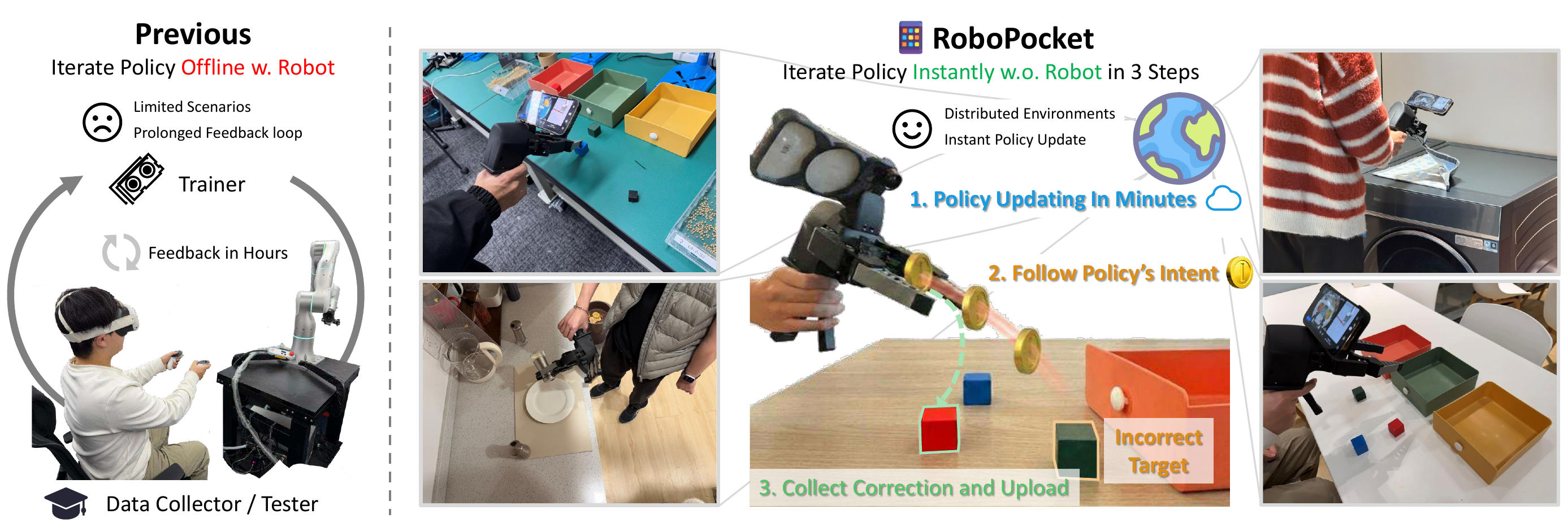}
\caption{Unlike previous workflows (left) relying on prolonged offline feedback loops with physical robots, \textbf{RoboPocket} (right) enables instant policy updates in distributed environments using consumer smartphones. By visualizing the policy's intent via AR Visual Foresight, users can proactively identify weaknesses and provide corrective data that refines the policy in minutes.}
\label{fig:teaser}
\vspace{-0.5cm}
\end{figure}

We introduce \textbf{RoboPocket}, a portable system that resolves this trade-off by turning consumer smartphones into closed-loop interfaces for robot-free policy iteration (see \cref{fig:teaser}). By leveraging the smartphone's integrated display, on-device compute, and low-latency communication, RoboPocket transforms data collection from passive recording into computationally guided data generation.

RoboPocket provides two forms of feedback. Real-time SLAM stability and kinematic feasibility checks help users detect invalid demonstrations during collection. More importantly, \textbf{AR Visual Foresight} overlays the remotely inferred policy trajectory onto the real scene~\cite{newbury2022visualizing}, allowing users to inspect intended motions and collect targeted corrections before robot deployment.

Coupled with asynchronous finetuning and model synchronization, this feedback creates a DAgger-inspired robot-free correction loop: users inspect policy intent, collect corrections, and query an updated policy within minutes. RoboPocket is a systems contribution that integrates these capabilities into a portable policy-iteration workflow.

We evaluate RoboPocket across diverse real-world manipulation tasks, including long-horizon sorting, pouring, deformable towel folding, and bimanual snack bagging. Our experiments demonstrate clear data-scaling trends and show that AR-guided online correction improves sample efficiency by up to \textbf{2$\times$} compared with scaling offline demonstrations. Furthermore, in distributed in-the-wild experiments, multiple users refine a shared policy across different environments with as few as 12 interactive corrections per user, demonstrating the scalability of robot-free policy iteration.

In summary, our key contributions are:
\textbf{(1) RoboPocket Data Collection System}: A smartphone-based system that transforms in-the-wild data collection into an active, computationally guided workflow, using real-time SLAM and kinematic feedback to ensure data validity.
\textbf{(2) Robot-Free Instant Policy Iteration}: A novel interactive paradigm powered by AR Visual Foresight and asynchronous finetuning. It allows users to visualize policy intent and provide targeted corrections without deploying physical robots.

\section{Related Work}

\textbf{Data Collection for Manipulation.}
Teleoperation remains the standard for collecting high-quality manipulation data. While master-slave systems (e.g., ALOHA~\cite{zhao2023learning, fu2024mobile}, GELLO~\cite{wu2024gello}) provide precision, and vision-based interfaces~\cite{qin2023anyteleop, cheng2024open, ding2025bunny, xue2025reactive} replace bulky hardware, both remain tied to physical robots, hindering large-scale in-the-wild collection. To overcome this bottleneck, robot-free interfaces utilize wearable systems~\cite{fang2024airexo, fang2025airexo, fang2025dexop} or handheld grippers~\cite{shafiullah2023bringing, chi2024universal, liu2024fastumi, liu2025forcemimic, chen2025arcap, wang2024dexcap, generalist2025gen0, rdt2} for portable collection, with notably UMI~\cite{chi2024universal} and its variants~\cite{liu2024fastumi, liu2025forcemimic} leveraging SLAM for scalable tracking. However, these systems are fundamentally \textbf{open-loop}: they record demos but lack feedback on data validity or policy weaknesses. In contrast, RoboPocket transforms handheld collection into a \textbf{closed-loop} engine, providing real-time validation and policy-aware guidance in the wild.

\textbf{Interactive Policy Learning and Correction.}
While Behavior Cloning scales well, it suffers from covariate shift~\cite{ross2011reduction}. Interactive methods like DAgger~\cite{ross2011reduction, kelly2019hg, zhang2016query, menda2019ensembledagger, hoque2021thriftydagger} and real-world RL~\cite{kalashnikov2018scalable, nair2020awac, nakamoto2023cal, pan2026sop, intelligence2025pi, luo2024serl, lei2025rl} mitigate this by collecting corrective data from policy-induced states, but executing imperfect policies on physical robots severely limits safety, frequency, and scalability. Moreover, existing shared autonomy~\cite{hagenow2021corrective, reddy2018shared} and interactive imitation systems~\cite{mandlekar2020human, liu2025robot, liu2024model} offer limited intent visibility, typically requiring users to intervene only \textit{after} failures occur. RoboPocket instead preempts failures: by visualizing predicted trajectories via AR, users can observe intent and collect targeted corrections without robot deployment, achieving \textbf{robot-free policy iteration} that merges the scalability of handheld collection with the targeted feedback of interactive learning.

\section{RoboPocket System Design}
\label{sec:system-design}

\begin{figure}[t]
\centering
\includegraphics[width=\linewidth]{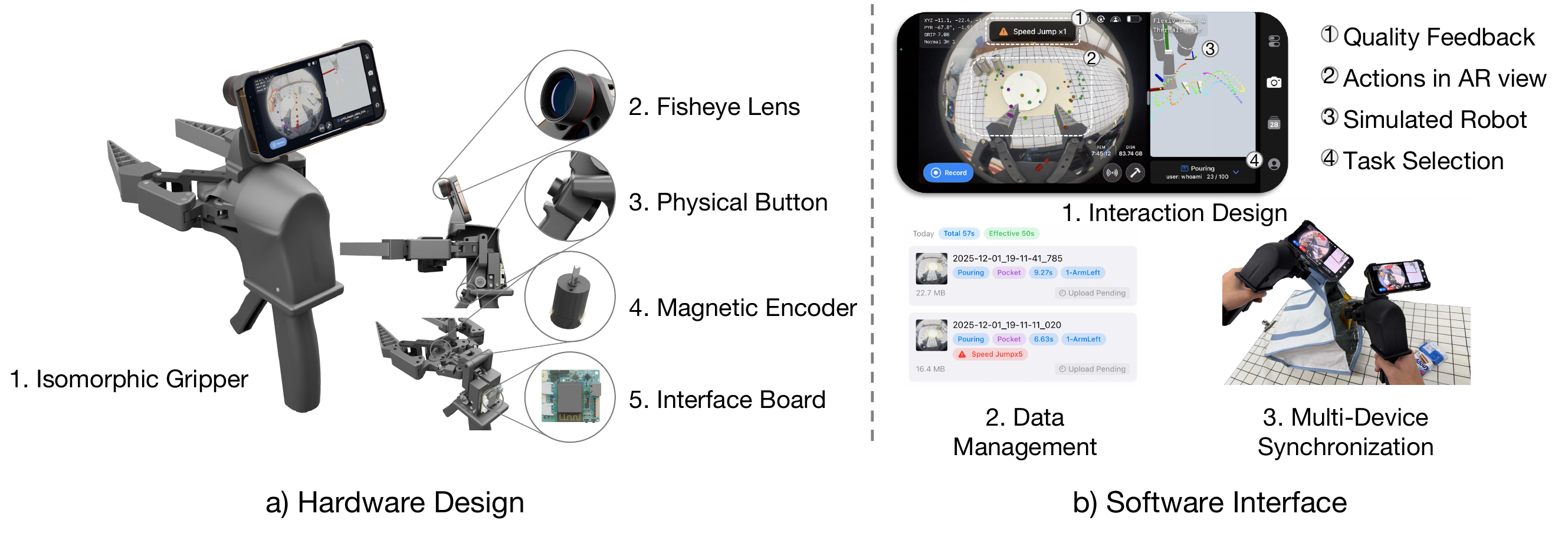}
\caption{\textbf{RoboPocket System Design.}
(a) \textbf{Hardware Design:} RoboPocket combines a consumer smartphone with a low-cost, 3D-printed adaptive gripper that is isomorphic to the Robotiq 2F-85. A fisheye lens expands the visual field of view, while an external gripper-width sensor captures proprioceptive information.
(b) \textbf{Software Interface:} The iOS application provides real-time feedback for tracking quality and kinematic validity, visualizes recorded or predicted trajectories in AR, and synchronizes multiple devices for bimanual collection.}
\label{fig:system_design}
\vspace{-0.5cm}
\end{figure}

As shown in \cref{fig:system_design}, RoboPocket combines a smartphone-based handheld collector, an isomorphic adaptive gripper, additional sensors, and an iOS interface. Its design supports \textbf{real-time interaction} during collection, \textbf{hardware isomorphism} for transfer to the robot, and \textbf{sensory completeness} for policy learning.

\paragraph{Hardware Architecture}
The smartphone runs visual-inertial tracking, kinematic checks, AR rendering, and network communication on device, enabling immediate collection feedback and low-latency interaction with remote policy inference.

The handheld gripper matches the geometry and approximates the adaptive finger behavior of the robot's Robotiq 2F-85, reducing the embodiment gap in robot-free demonstrations. Mechanical implementation details are provided in the appendix. A fisheye lens expands the wrist-view field of view to include gripper--object interactions and the surrounding scene. An external encoder streams gripper width to the phone, complementing the visual and pose observations. Sensor and communication details are provided in the appendix.

\paragraph{Software Architecture}
The software of RoboPocket turns on-device computation into real-time data-quality feedback. During collection, the iOS application monitors both tracking reliability and kinematic feasibility. It detects unstable visual-inertial tracking and checks whether the recorded gripper motion can be mapped to feasible robot joint configurations. When invalid or risky states are detected, the interface immediately alerts the user through visual feedback, allowing them to adjust the motion before unusable data is stored. After each demonstration, AR trajectory replay helps users inspect SLAM drift and motion coherence directly in the collection scene, complementing the online validity checks.

RoboPocket also supports synchronized multi-device collection for bimanual manipulation. Multiple phones establish a shared spatial coordinate frame and align their timestamps, so images, trajectories, and gripper states from different devices can be fused into the same world frame. This synchronization allows the same smartphone-based interface to scale from single-arm tasks to coordinated dual-arm data collection.

\section{Robot-Free Instant Policy Iteration}
\label{sec:policy_iteration}

\begin{figure}[t]
    \centering
    \includegraphics[width=\textwidth]{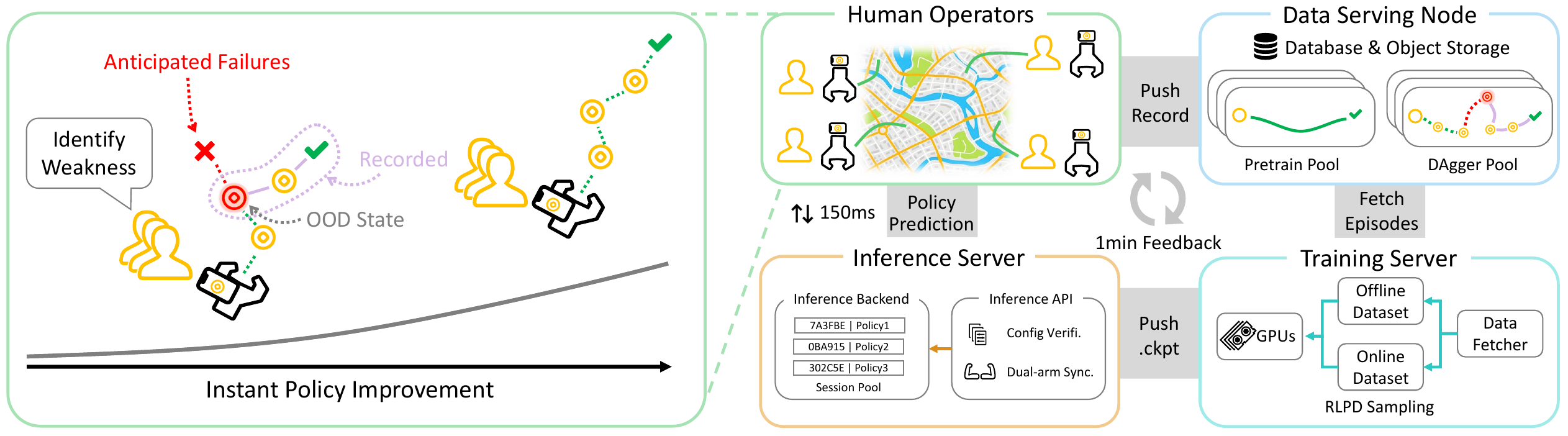}
    \caption{\textbf{Overview of Robot-Free Instant Policy Iteration.}
(Left) Using AR Visual Foresight, the user identifies policy weaknesses (OOD states) and proactive failures in the real world, which improves the policy instantly.
(Right) Collected corrective data is immediately streamed to the \textit{Data Serving Node}. The \textit{Training Server} performs online finetuning using RLPD sampling~\cite{ball2023efficient} and pushes updated weights to the \textit{Inference Server}. The improved policy predictions are streamed back to iPhone in real-time ($<150$ms), enabling continuous, robot-free policy improvement \textbf{in minutes}.}
\label{fig:in-the-wild}
\vspace{-0.5cm}
\end{figure}

Interactive imitation learning improves robustness by collecting corrective data from states induced by the current policy~\cite{ross2011reduction}. However, conventional approaches require executing imperfect policies on physical robots, which is costly, unsafe, and difficult to scale across diverse environments. RoboPocket removes this dependency by making policy intent visible directly through the handheld collector. As shown in \cref{fig:in-the-wild}, users can query the current policy in the real world, observe its predicted behavior through AR, and collect corrections in the states where the policy is likely to fail. This creates a robot-free approximation of on-policy data collection: the policy is never executed by a robot, but its predicted actions still guide users toward policy-relevant state distributions.

\subsection{Robot-Free Policy Feedback}

RoboPocket implements policy feedback through an edge--cloud inference pipeline. The smartphone acts as a lightweight client that captures observations, tracks the handheld gripper pose, and renders feedback to the user. The policy runs on a cloud-side inference server, which receives streamed observations and returns the predicted action trajectory in real time. This design allows computationally heavy policy inference to run on a GPU server while preserving an interactive user experience on the mobile device.

\begin{wrapfigure}{r}{0.37\linewidth}
\centering
\includegraphics[width=\linewidth]{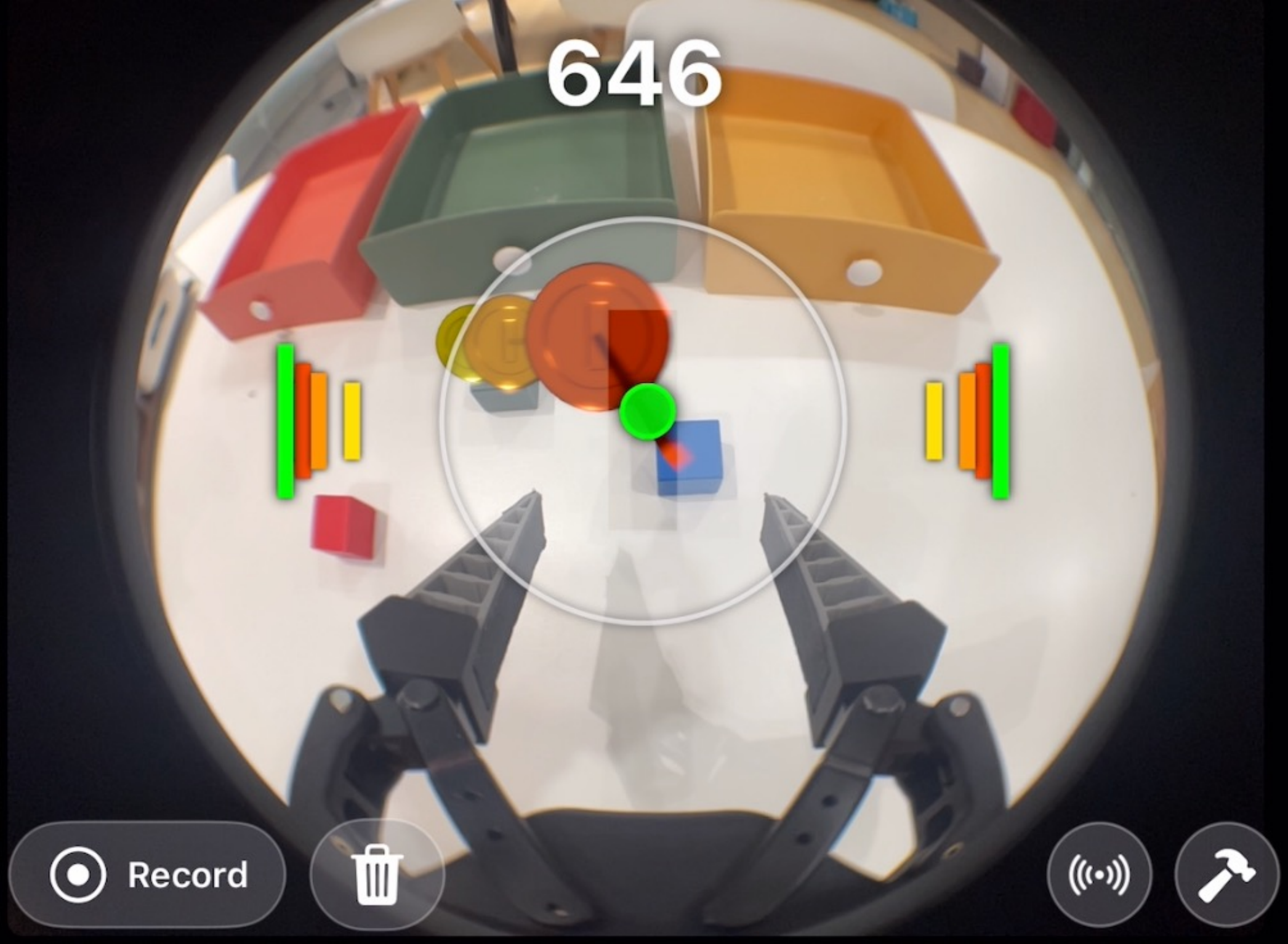}
\caption{\textbf{AR Visual Foresight.} Oriented virtual coins visualize predicted end-effector poses.}
\label{fig:ar-foresight}
\vspace{-0.4cm}
\end{wrapfigure}

The returned trajectory is visualized in the phone's camera view as \textbf{AR Visual Foresight}. Each predicted end-effector pose is rendered as an oriented virtual coin, conveying both translation and rotation. Coin orientations indicate roll, while a dot on the user coin indicates pitch/yaw alignment with the next predicted action: a centered dot denotes alignment, and vertical or horizontal offsets indicate pitch or yaw misalignment, respectively. Users can inspect these cues and collect a corrective demonstration when the intended motion appears incorrect or likely to fail.

RoboPocket further supports proactive intervention during collection. Users can trigger a new policy query at any time, inspect the updated prediction, and decide whether to follow, correct, or explore around the suggested trajectory. Through repeated interaction with the current policy, users naturally discover weak regions of the policy and collect targeted demonstrations that are more informative than passive, open-loop recordings.

\subsection{Instant Policy Iteration}

To turn robot-free feedback into policy improvement, RoboPocket couples AR-guided correction with asynchronous online finetuning. During collection, each corrective trajectory is uploaded to the backend as soon as it is recorded. The training server continuously incorporates the newly collected corrections while retaining access to the original demonstration dataset, allowing the policy to adapt to recent failure cases without discarding the broad coverage of the offline data.

Updated policy checkpoints are periodically synchronized to the inference server. As a result, the next AR Visual Foresight query reflects the improved policy rather than a fixed pretrained model. This closes the loop between policy evaluation, corrective data collection, and policy update: users identify a failure in AR, collect targeted data, and observe the updated policy behavior within minutes. RoboPocket therefore realizes a DAgger-inspired robot-free correction loop, where policy-aware corrective data can be collected in the wild without requiring physical robot rollouts. The formulation and implementation details are shown in the appendix.

\begin{figure}[t]
    \centering
    \includegraphics[width=\linewidth]{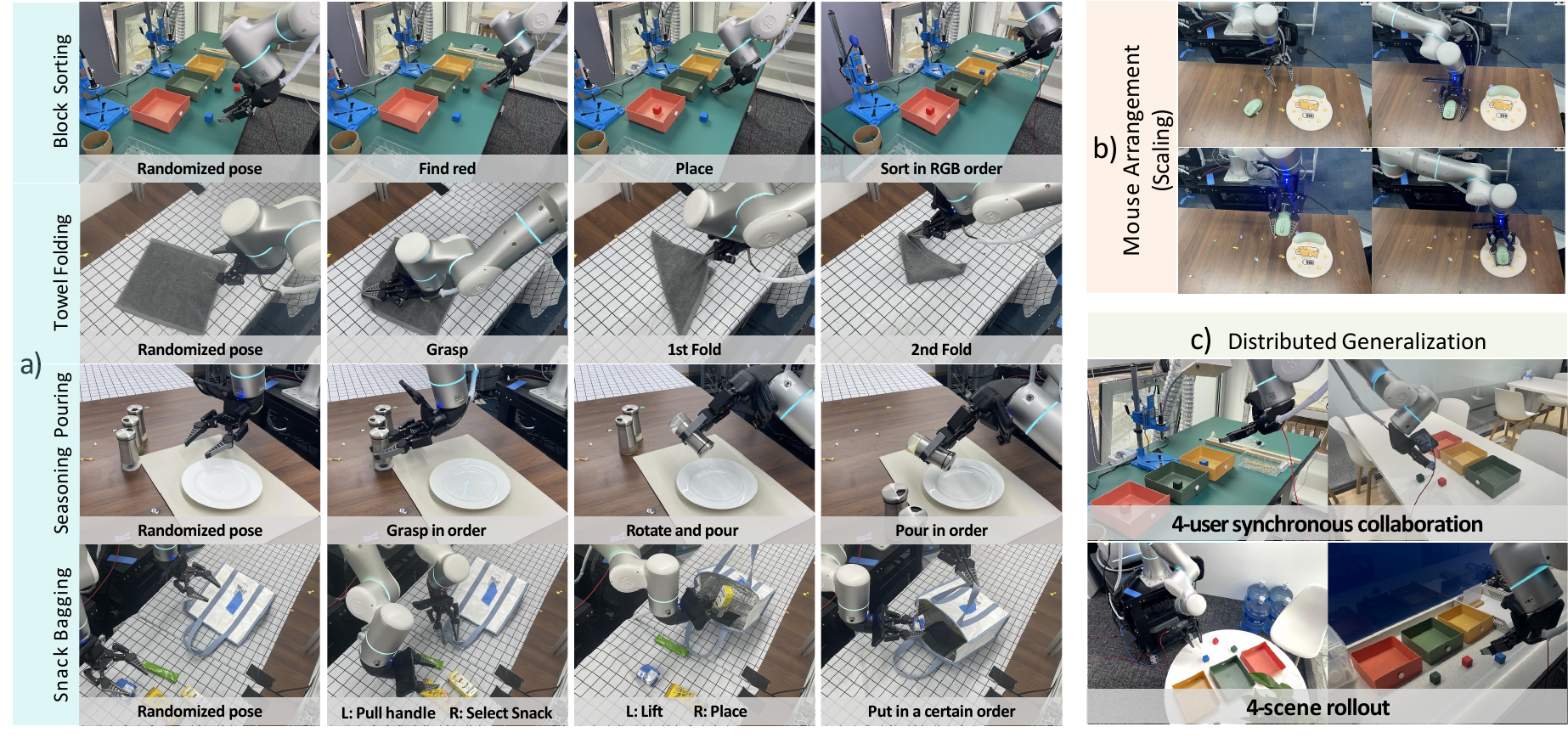}
    \caption{
        \textbf{Evaluation Tasks.}
        \textbf{a)} We evaluate our method on four manipulation tasks: Block Sorting, Seasoning Pouring, Towel Folding, and Snack Bagging.
        \textbf{b)} Mouse Arrangement validates data fidelity and scaling; the four images are keyframes from a single rollout. Evaluation uses diverse novel object pairs, scenes, and initial poses.
        \textbf{c)} To assess distributed adaptation, 4 users collect data and perform in-the-wild policy iteration across distinct environments for the Block Sorting task.
    }
    \label{fig:tasks}
    \vspace{-0.5cm}
\end{figure}

\section{Experiments}
\label{sec:experiments}

We evaluate RoboPocket along three axes: (1) \textbf{whether it collects valid robot learning data}, (2) \textbf{whether robot-free instant policy iteration improves data efficiency beyond offline scaling}, and (3) \textbf{whether the same workflow scales to distributed in-the-wild settings}. As shown in \cref{fig:tasks}, our experiments include a data-scaling validation task, four real-world manipulation tasks with various challenges, and a distributed multi-user policy-iteration study. All robot evaluations use Flexiv Rizon 4~\cite{flexiv_rizon4} arms with Robotiq 2F-85~\cite{robotiq_2f85} grippers. Policies are trained with Diffusion Policy~\cite{chi2025diffusion} using CLIP~\cite{radford2021learning} or DINOv2~\cite{oquab2023dinov2} visual encoders. Please refer to the appendix for detailed setups.

\subsection{Data Fidelity and Scaling Validation}
\label{subsec:system_verification}

We verify trajectory fidelity, collection validity, and policy scaling with RoboPocket data.

\begin{wrapfigure}{r}{0.5\linewidth}
\centering
\includegraphics[width=\linewidth]{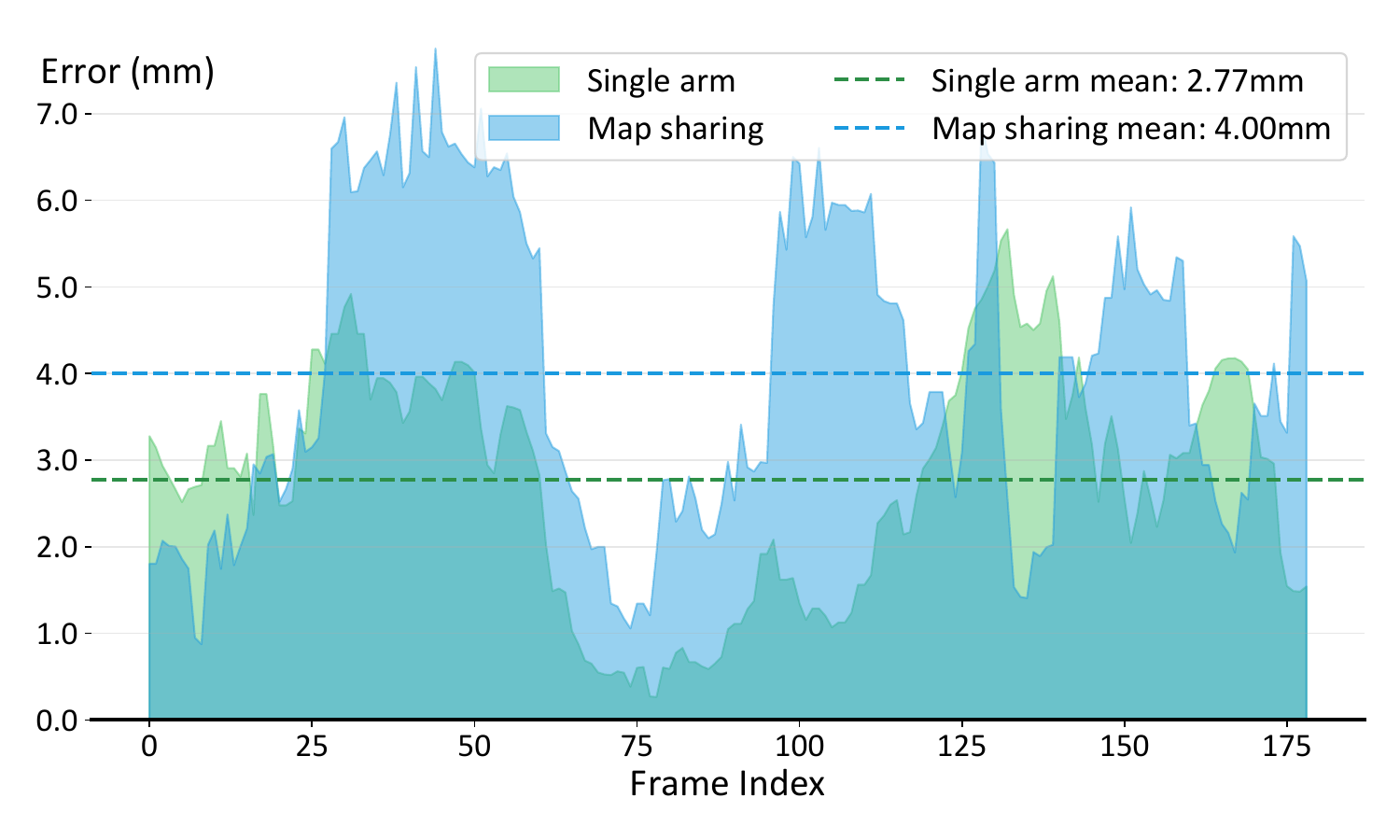}
\caption{\textbf{RoboPocket Localization Accuracy.} We measure cumulative 3D Euclidean trajectory error against robot kinematic ground truth.}
\label{fig:tracking_accuracy}
\vspace{-0.4cm}
\end{wrapfigure}

\textbf{Trajectory fidelity.}
We rigidly attach RoboPocket to the robot end-effector and compare tracking with robot kinematic ground truth. As shown in \cref{fig:tracking_accuracy}, single-device position and rotation errors are 2.8mm and 0.4$^\circ$, compared with 6.1mm and 3.5$^\circ$ for the UMI inertial-monocular SLAM pipeline~\cite{chi2024universal}. Under the same setup, ARKit without the fisheye lens yields 2.3mm and 0.4$^\circ$: the lens adds 0.5mm positional error with no measured increase in rotation error. Dual-device synchronization yields 4.0mm average position error (7.5mm peak) and 0.7$^\circ$ rotation error. The interface flags unstable tracking in low-texture scenes during collection.

\textbf{Collection efficiency and data validity.}
We further compare RoboPocket with a standard UMI-style handheld pipeline on a Seasoning Pouring collection task. For 10 demonstrations, UMI requires 8m34s for collection, 1m24s for transfer, and 9m12s for offline SLAM processing. RoboPocket removes the offline SLAM stage and only requires 3m51s for acquisition and 1m37s for transfer. The collected trajectories are also more stable: UMI produces significant position jumps in 2 out of 9 successful trials and acceleration spikes above $15m/s^2$ in every trajectory, while RoboPocket yields no position jumps and maintains physically plausible acceleration profiles.

To isolate the effect of active verification, we collect two 50-trajectory sessions in a low-texture environment, with and without RoboPocket's real-time feedback. Active verification combines tracking-state feedback, kinematic feasibility checks, and AR trajectory visualization. It reduces abnormal trajectories from 30\% to 0\%, where abnormality is defined as more than 1m localization drift or more than 5cm trajectory jumps. These results show that RoboPocket not only records trajectories accurately, but also helps users avoid unusable data during collection.

\begin{figure}[t]
\centering
\includegraphics[width=0.9\linewidth]{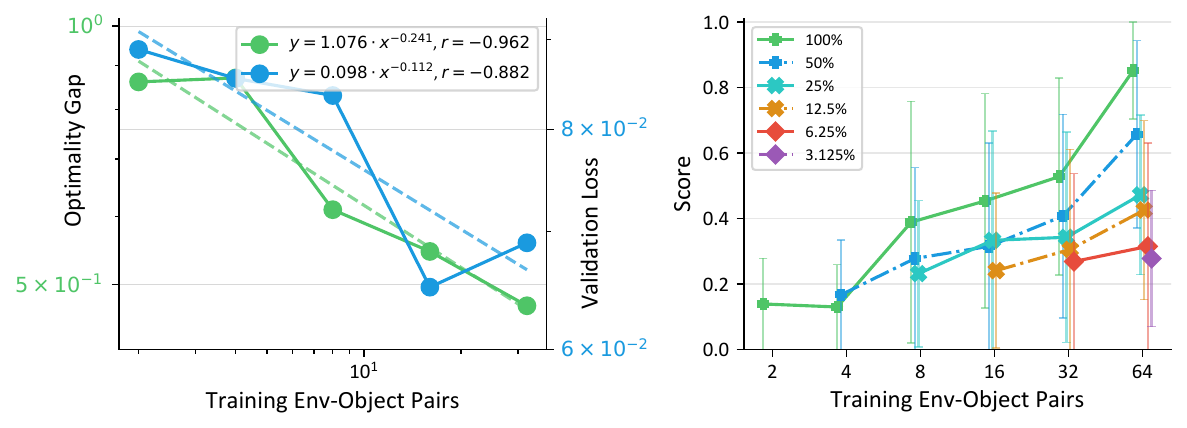}
\caption{\textbf{Data Scaling Trend Validation.}
On the Mouse Arrangement task, policy performance shows strong power-law correlations with environment-object diversity ($r=-0.962$ and $r=-0.882$), and consistently improves with more training data.}
\label{fig:scaling_laws}
\vspace{-0.3cm}
\end{figure}

\textbf{Scaling behavior.}
On 1,600 Mouse Arrangement demonstrations spanning 64 environment-object pairs, OOD policy performance improves with environment-object diversity (\cref{fig:scaling_laws}), consistent with prior data-scaling observations~\cite{hu2024data}. This experiment validates the collected data; the following experiments evaluate the benefit of policy-aware correction.

\begin{table}[!t]
\centering
\begin{minipage}[t]{0.49\linewidth}
\vspace{0pt}
\centering
\caption{\textbf{DAgger comparison on Block Sorting.} 100 base demonstrations, 50 corrections, and 20 evaluation trials per policy.}
\label{tab:dagger-compare}
\vspace{5pt}
\scriptsize\setlength{\tabcolsep}{2pt}
\begin{adjustbox}{max width=\linewidth}
\begin{tabular}{lccc}
\toprule
 & \shortstack{Robot Offline\\DAgger} & \shortstack{Robot Online\\DAgger} & \shortstack{Instant PI\\(Ours)} \\
\midrule
Normalized score & 0.80 & 0.88 & 0.91 \\
Collection time (s) & -- & 91.5 & 55.7 \\
\bottomrule
\end{tabular}
\end{adjustbox}
\end{minipage}\hfill
\begin{minipage}[t]{0.49\linewidth}
\vspace{0pt}
\centering
\caption{\textbf{Single- vs. multi-scene corrections.} Normalized scores; Multi uses 12 corrections from each of four scenes.}
\label{tab:scene-diversity}
\vspace{6pt}
\scriptsize\setlength{\tabcolsep}{2pt}
\begin{adjustbox}{max width=\linewidth}
\begin{tabular}{lcccc}
\toprule
 & Base & +12 Single & +48 Single & +48 Multi \\
\midrule
Scene 4 & 0.52 & 0.81 & 0.86 & 0.81 \\
Other scenes & 0.66 & 0.76 & 0.87 & 0.93 \\
\bottomrule
\end{tabular}
\end{adjustbox}
\end{minipage}
\end{table}

\subsection{Data Efficiency of Robot-Free Policy Iteration}
\label{subsec:beyond_scaling}

While offline data scaling improves imitation learning, it remains inefficient when additional demonstrations do not target the states where the current policy fails. We therefore evaluate whether \textit{Robot-Free Instant Policy Iteration} can accelerate scaling by collecting policy-aware corrective data. As shown in \cref{fig:tasks}, we test four tasks with complementary challenges: \textbf{Block Sorting}, a long-horizon sequential task; \textbf{Seasoning Pouring}, which requires large wrist rotations and precise repositioning; \textbf{Towel Folding}, which involves deformable-object perception and manipulation; and \textbf{Snack Bagging}, a bimanual coordination task. We compare \textbf{IL + Instant PI} with three baselines: (1) \textbf{IL Only}, trained on fixed offline datasets of size $N=100,200,300$; (2) \textbf{IL + Manual PI}, where an expert inspects robot failure videos and collects targeted corrections; and (3) \textbf{IL + Offline PI}, where users collect corrections using RoboPocket's AR feedback but with a fixed pretrained policy.

\begin{figure}[t]
\centering
\includegraphics[width=\linewidth]{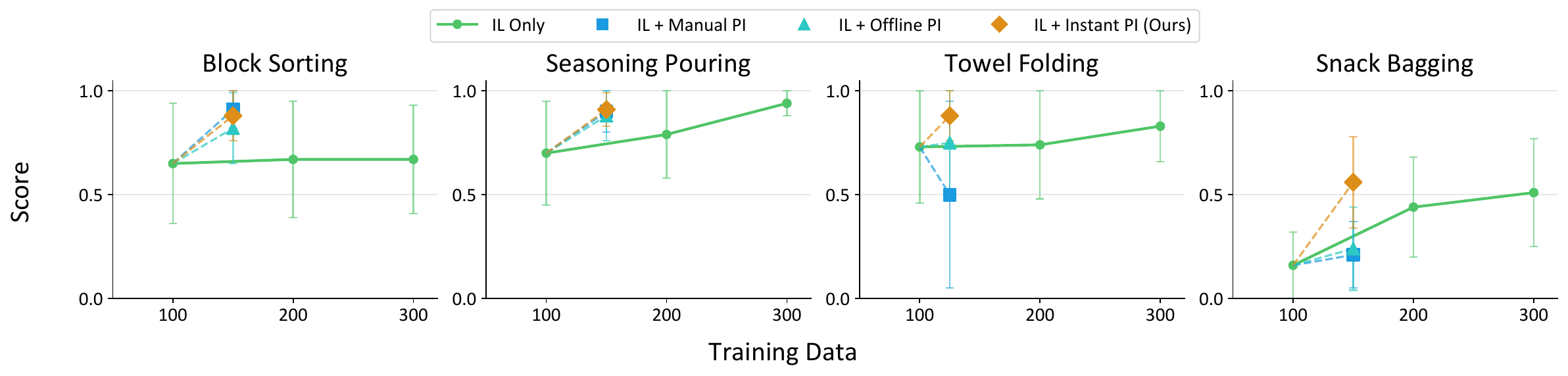}
\caption{\textbf{Performance Comparison.}
Across four manipulation tasks, \textbf{IL + Instant PI} improves data efficiency by up to \textbf{2$\times$} over pure offline scaling and achieves performance comparable to manual policy iteration without requiring physical robot deployment.}
\label{fig:main_results}
\vspace{-0.2cm}
\end{figure}

\textbf{Instant policy iteration improves scaling efficiency.}
As shown in \cref{fig:main_results}, pure IL improves with more offline data, but its gains are limited by covariate shift and task-specific failure modes: wrong execution order in Block Sorting, poor recovery after large rotations in Seasoning Pouring, incorrect semantic grasp points in Towel Folding, and grasping or occlusion failures in bimanual Snack Bagging. By exposing these failures through AR Visual Foresight, RoboPocket guides users to collect corrections near policy-relevant states. Across all four tasks, \textbf{IL + Instant PI} reaches or exceeds the performance of substantially larger offline datasets, achieving up to \textbf{2$\times$} higher data efficiency than pure offline scaling.

\textbf{Online feedback improves correction quality.}
Compared with \textbf{IL + Offline PI}, instant policy iteration gives users feedback from the updated policy during collection rather than from a fixed pretrained model. This is especially important when corrective data can be noisy or task-dependent. For Seasoning Pouring, \textbf{IL + Instant PI} achieves much lower variance than \textbf{IL + Offline PI} ($0.08$ vs. $0.30$), indicating more stable corrections. For Towel Folding, manual correction can even reduce performance ($0.73 \rightarrow 0.50$), while \textbf{IL + Instant PI} yields stable improvement ($0.88$), suggesting that seeing the current policy's intent helps users avoid collecting misleading recovery data. Overall, RoboPocket provides policy-aware guidance comparable to expert manual intervention, but without requiring robot rollouts.

\textbf{Comparison with robot-in-the-loop DAgger.}
We additionally compare Instant PI with robot-in-the-loop DAgger~\cite{ross2011reduction} on Block Sorting. All three variants start from 100 base demonstrations, collect 50 corrections, and are evaluated over 20 robot trials per policy. As shown in \cref{tab:dagger-compare}, Instant PI attains a normalized score of 0.91, compared with 0.88 for Robot Online DAgger, while reducing average collection time per corrective demonstration from 91.5s to 55.7s (39\%). This supports the practical benefit of querying and correcting the policy through a handheld interface without robot execution.

\textbf{Proactive intervention reduces correction cost.}
We isolate proactive intervention on Block Sorting, starting from 100 base demonstrations and collecting 50 corrections per variant with the same online finetuning setup. In the baseline, users start from random states and follow the AR trajectory until the policy succeeds or deviates, intervening to record recovery data when needed. With proactive intervention, users can re-query while exploring, identify weak state regions, and collect corrections earlier. Both variants achieve comparable task scores (5.25 vs. 5.30), while proactive intervention reduces average time per effective correction from 75.7s to \textbf{39.1s}.

\subsection{Distributed Policy Iteration across Scenes}
\label{subsec:scalable_generalizable}

A key advantage of robot-free policy iteration is that correction data can be collected outside the robot lab and across multiple environments. To evaluate this distributed setting, four users perform Block Sorting data collection in four distinct rooms. We first collect 100 demonstrations in total, with 25 demonstrations per scene, to train a base policy. Each user then performs \textit{Robot-Free Instant Policy Iteration} in their own environment and collects only 12 corrective demonstrations.

\begin{figure}[t]
\centering
\includegraphics[width=\linewidth]{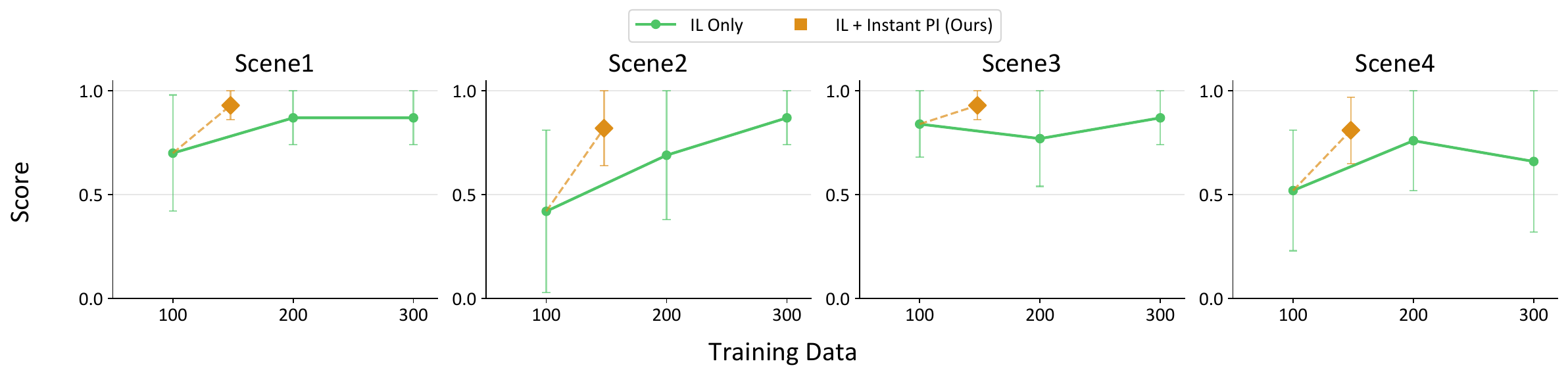}
\caption{\textbf{Distributed Policy Iteration.}
RoboPocket enables users to improve policy performance across four distinct environments with only 12 interactive corrections per scene.}
\label{fig:distributed_results}
\vspace{-0.3cm}
\end{figure}

As shown in \cref{fig:distributed_results}, 12 corrections per scene improve performance in challenging environments, including Scene 2 ($0.42 \rightarrow 0.82$) and Scene 4 ($0.52 \rightarrow 0.81$).

\textbf{Correction diversity improves performance across scenes.}
To separate correction quantity from scene diversity, we compare 12 or 48 corrections collected only in Scene 4 with 48 multi-scene corrections (12 per scene). At the same 48-correction budget (\cref{tab:scene-diversity}), multi-scene collection yields a score of 0.93 on the other scenes, compared with 0.87 for single-scene collection, while achieving 0.81 versus 0.86 on Scene 4. Thus, distributed corrections retain substantial target-scene adaptation while improving performance across the other scenes. These other scenes also contribute corrections in the multi-scene setting, so the comparison measures adaptation across scenes rather than transfer to wholly unseen environments.

\section{Conclusion and Limitations}
\label{sec:conclusion}

RoboPocket integrates AR Visual Foresight, remote inference, and asynchronous finetuning into a smartphone-based system for robot-free policy iteration. Experiments demonstrate improved data efficiency, reduced correction cost relative to robot-in-the-loop DAgger, and adaptation across distributed collection scenes.

\textbf{Limitations.} AR Visual Foresight exposes motion-level errors, including subtle position or orientation misalignment, but kinematic trajectories alone cannot reliably reveal failures governed by force, friction, or compliance. The handheld gripper also only approximates robot contact dynamics, and its parallel-jaw design limits dexterous in-hand manipulation. Robot-free corrections are guided by human motion and policy predictions rather than actual robot rollouts, so their state distribution can differ from that induced by the deployed policy. The current system requires network access for policy feedback, and the handheld rig may cause fatigue during prolonged use. Future work includes force or tactile visualization, on-device inference, and lighter interfaces such as AR glasses.

\acknowledgments{
We would like to thank the reviewers for their constructive comments. We thank Yongkai Fan for inspiring the hardware design used in our experiments. We also thank Yuan Fang, Nan Min, Baijie Shao, Yibo Shen, Jianghua Shi, Jiahao Zhang, and Yanwen Zou for participating in the user study.

This work was supported by the Fundamental and Interdisciplinary Disciplines Breakthrough Plan of the Ministry of Education of China, the Science and Technology Major Project of Jiangsu Province (No. SBG2024041), Shanghai Artificial Intelligence Laboratory, and XPLORER PRIZE grants.
}
\bibliography{references}

\newpage

\appendix

\section{RoboPocket System Design Details}

This section provides additional implementation details of the RoboPocket hardware and software system described in \cref{sec:system-design}.

\subsection{Isomorphic Adaptive Gripper Design}

RoboPocket is designed to match the geometry and passive behavior of the Robotiq 2F-85 gripper used in our robot setup. Unlike generic handheld parallel grippers, our collector follows the visual geometry and grasping configuration of the target robot gripper, reducing the embodiment gap between robot-free demonstrations and robot execution.

To better reproduce contact-rich interactions, the handheld gripper approximates the adaptive finger behavior of the Robotiq 2F-85. We integrate pre-compressed torsion springs into the distal finger joints to emulate the passive degree of freedom of the real gripper. This allows the handheld device to exhibit passive finger deformation during compliant grasping or unintended surface contact, so the collected trajectories better reflect the physical behavior of the deployed gripper.

The gripper also uses a leverage-based linkage mechanism to amplify human finger input. This makes it easier for users to apply and maintain sufficient grasping force during long data collection sessions. The main structure is 3D printed using standard FDM processes, and the full handheld assembly has a bill of materials cost of approximately $\$70$, excluding the smartphone.

\subsection{Augmented Sensing Modules}

Although modern smartphones provide high-quality cameras, IMUs, and visual-inertial tracking, they do not directly provide all sensing channels needed for robot policy learning. RoboPocket therefore augments the phone with two additional sensing components.

First, we attach an off-the-shelf fisheye lens through a custom 3D-printed mount. This expands the effective field of view of the wrist-mounted camera, allowing the observation to capture both the gripper-object interaction and a larger portion of the surrounding workspace. This is important for manipulation tasks where object context, task progress, and gripper pose must be inferred from the same wrist-view observation.

Adapting the collector to a different iPhone model requires a matching 3D-printed shell to maintain consistent mounting and fisheye-lens alignment with the camera module. This mechanical adaptation does not establish identical tracking or policy performance across phone models; a controlled comparison remains outside the current evaluation.

Second, RoboPocket measures gripper width using an external magnetic encoder. The encoder signal is transmitted to an ESP32-based interface board through a 1 Mbps RS485 bus, achieving an angular resolution of $0.088^{\circ}$ at 30Hz. The ESP32 streams the gripper-width state to the iPhone via BLE GATT. The firmware follows a lightweight publisher-subscriber pattern, which keeps the gripper state synchronized with visual and pose observations while leaving the interface extensible to additional sensors.

\subsection{Real-Time Data Verification}

RoboPocket performs online verification to identify unusable or risky data during collection rather than after offline post-processing. The verification module monitors both tracking reliability and kinematic feasibility.

For tracking reliability, the system monitors visual-inertial odometry quality using indicators such as feature density and sudden pose or velocity jumps. Frames with unstable tracking are marked as invalid, and the interface immediately notifies the user through visual feedback. This allows users to adjust the scene, motion speed, or viewpoint before severe drift corrupts the trajectory.

For kinematic feasibility, RoboPocket runs an on-device inverse-kinematics solver to map the handheld gripper motion to the robot joint space. We use a damped least-squares Jacobian solver~\cite{nakamura_inverse_1986} and check for infeasible configurations such as joint-limit violations or near-singular motions. When the recorded pose is difficult or impossible for the robot to execute, the interface provides real-time warnings so that the user can correct the motion during collection.

\subsection{AR Trajectory Replay}

In addition to algorithmic checks, RoboPocket provides AR trajectory replay for in-situ trajectory inspection. After a demonstration, the recorded end-effector trajectory is rendered back into the real scene through the phone display. This allows the user to verify whether the reconstructed trajectory aligns with the physical motion and whether the demonstration completes the intended task. In practice, this replay function helps reveal SLAM drift, discontinuous pose jumps, unstable grasps, and task-level failures immediately after collection.

\subsection{Multi-Device Spatiotemporal Synchronization}

For bimanual manipulation, RoboPocket synchronizes multiple smartphone collectors in both space and time. Spatial alignment is achieved through a peer-to-peer map sharing protocol based on ARKit~\cite{arkit}. Devices exchange world maps and establish a shared coordinate frame, allowing trajectories from different phones to be represented in the same world frame.

Temporal alignment is handled by a lightweight network synchronization protocol that aligns device clocks with approximately 5ms precision. Each sensor packet, including images, poses, and gripper-width measurements, is timestamped on device and later aligned across devices. This ensures that demonstrations collected from multiple handheld interfaces can be fused into synchronized multi-arm trajectories for policy learning.

\section{Robot-Free Instant Policy Iteration Details}

This section provides additional formulation and implementation details for the Robot-Free Instant Policy Iteration workflow described in \cref{sec:policy_iteration}.

\subsection{Connection to Interactive Imitation Learning}

We formulate a manipulation task as a Markov Decision Process (MDP) $(\mathcal{S}, \mathcal{A}, \mathcal{P}, \mathcal{R}, \gamma)$. Standard imitation learning trains a policy $\pi_\theta(\mathbf{a}_t \mid \mathbf{s}_t)$ from a static demonstration dataset $\mathcal{D}_{\mathrm{demo}}=\{(\mathbf{s}_t,\mathbf{a}_t)\}$, but the learned policy may encounter out-of-distribution states due to compounding errors. Let $d_\pi$ denote the state distribution induced by policy $\pi$. The target objective is to minimize the imitation loss under the policy-induced distribution:
\begin{equation}
J(\pi) = \mathbb{E}_{\mathbf{s} \sim d_\pi} [\ell(\pi(\mathbf{s}), \pi^*(\mathbf{s}))],
\end{equation}
where $\pi^*$ denotes the expert policy.

Interactive imitation learning methods such as DAgger~\cite{ross2011reduction} address this issue by aggregating corrective data from the state distributions induced by intermediate policies. Let $d_{\pi_i}$ be the state distribution induced by policy $\pi_i$ at iteration $i$. After $N$ iterations, an idealized aggregate-distribution objective is
\begin{equation}
\bar{d}_N(\mathbf{s}) =
\frac{1}{N}\sum_{i=1}^{N} d_{\pi_i}(\mathbf{s}),
\quad
\hat{\pi}_{N+1}
=
\arg\min_{\pi \in \Pi}
\mathbb{E}_{\mathbf{s} \sim \bar{d}_N}
[\ell(\pi(\mathbf{s}), \pi^*(\mathbf{s}))],
\end{equation}
where $\Pi$ is the policy class. In practice, the policy is fitted to the aggregated labeled data.

RoboPocket follows the motivation of policy-aware data aggregation, but its corrections are collected from a human-mediated distribution $\tilde{d}_{\pi_i}$: users follow or probe predicted trajectories with the handheld collector and demonstrate corrections where needed. Since human motion and robot dynamics differ, $\tilde{d}_{\pi_i}$ need not equal $d_{\pi_i}$. Thus, RoboPocket provides a DAgger-inspired workflow whose effectiveness is evaluated empirically, without claiming the theoretical guarantees of robot-executed DAgger.

\subsection{Remote Inference Implementation}

RoboPocket uses an edge--cloud architecture for policy interaction. The smartphone acts as a lightweight client responsible for observation capture, pose tracking, user interaction, and AR visualization. Policy inference is offloaded to a GPU-equipped inference server. When a client starts a session, the server loads the corresponding policy checkpoint and configuration, maintains the policy state across queries, and returns predicted action trajectories to the phone.

During interaction, the smartphone streams the current observation to the inference server. The server predicts an action trajectory and sends it back to the client for AR visualization. By maintaining persistent model states and avoiding repeated model initialization, the system achieves real-time interaction with a round-trip latency below 150ms over standard Wi-Fi in our setup.

\subsection{AR Visual Foresight Rendering}

RoboPocket visualizes the predicted policy trajectory in the real scene through augmented reality. Since our camera uses a fisheye lens to expand the manipulation field of view, directly rendering virtual objects with a standard pinhole camera model would cause visible misalignment. We therefore apply distortion-aware rendering based on the calibrated camera intrinsics. In practice, the predicted trajectory is rendered as a sequence of virtual markers in the fisheye camera view, and a real-time vertex displacement procedure aligns the virtual trajectory with the distorted image.

Each marker is an oriented virtual coin representing a predicted end-effector pose. Relative coin orientations expose rotational motion, including roll. A dot on the coin representing the user's current pose indicates pitch/yaw alignment with the next action: centering indicates alignment, vertical displacement indicates pitch misalignment, and horizontal displacement indicates yaw misalignment. These cues let users inspect positional and rotational errors, such as a grasp pose that would collide with an object. They convey kinematic intent rather than predicted contact forces.

The AR interface is designed to make policy intent immediately interpretable. Users can follow the visualized trajectory when it appears correct, or provide a corrective demonstration when the prediction is wrong, uncertain, or likely to fail. When the device reaches the end of the current prediction horizon, the system captures a new observation and queries the policy again, allowing users to continuously interact with the current policy during data collection.

\subsection{Proactive Intervention}

In addition to automatic policy queries, RoboPocket provides a physical intervention button that allows users to request a new policy prediction at any time. This enables users to actively probe the current policy rather than passively following a fixed trajectory. For example, when the predicted trajectory approaches an ambiguous object configuration or an out-of-distribution state, the user can trigger a new query, inspect the updated prediction, and decide whether to collect a corrective demonstration.

This interaction mechanism changes correction from reactive safeguarding to proactive data collection. Instead of waiting for a robot to fail, users can explore the policy's weak regions in a robot-free setting and collect demonstrations that directly target those states.

\subsection{Asynchronous Online Finetuning}

The backend contains three services: an \textit{Inference Server}, a \textit{Data Serving Node}, and a \textit{Training Server}. The Inference Server handles real-time policy queries from smartphones. The Data Serving Node receives newly collected trajectories and manages the growing correction dataset. The Training Server monitors the Data Serving Node and continuously finetunes the policy when new corrective data arrives.

To balance adaptation and retention, we train on a mixture of the original offline dataset $\mathcal{D}_{\mathrm{demo}}$ and the newly collected correction dataset $\mathcal{D}_{\mathrm{on}}$. Following the intuition of replay-based policy updates such as RLPD~\cite{ball2023efficient}, each finetuning batch samples 50\% of data from $\mathcal{D}_{\mathrm{demo}}$ and 50\% from $\mathcal{D}_{\mathrm{on}}$. This allows the policy to quickly fit recent corrections while reducing catastrophic forgetting of the original demonstration distribution.

Updated checkpoints are periodically synchronized from the Training Server to the Inference Server. In our implementation, model weights are pushed every $N$ training steps. The next policy query from the smartphone therefore uses the latest available checkpoint, closing the loop between AR-guided correction, online finetuning, and updated policy feedback.

\section{User Study}

To validate whether RoboPocket enables users without task-specific training (\ie non-expert users) to identify policy weaknesses and collect correction data in a robot-free setting, we design a user study.

\subsection{Participants and Procedure}

\begin{figure}[h]
    \centering
    \includegraphics[width=\linewidth]{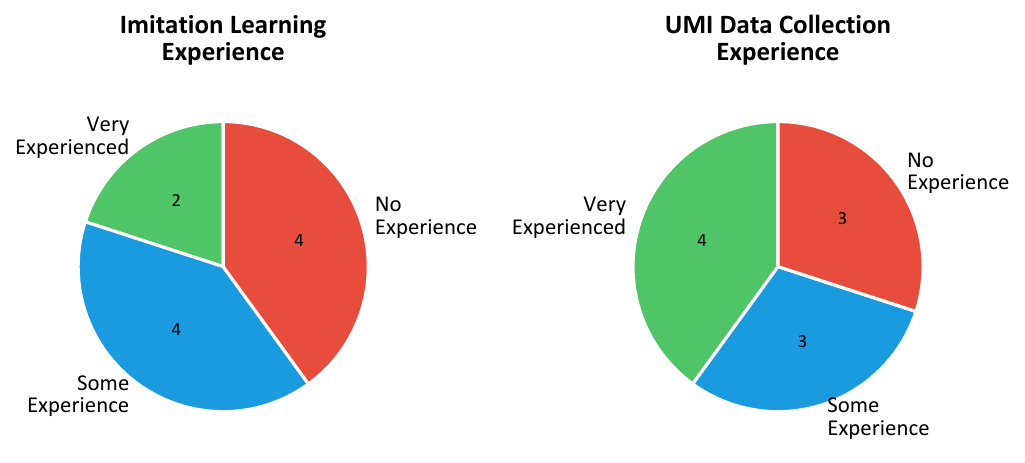}
    \caption{\textbf{Participant Demographics.} The distribution of volunteers based on their prior experience with imitation learning and UMI data collection.}
    \label{fig:user_demographics}
\end{figure}

We invite a group of volunteers with varying levels of experience in imitation learning and UMI-based data collection. The demographic distribution of the participants is illustrated in \cref{fig:user_demographics}.
The participants are invited to perform a Block Sorting task using \textit{Robot-Free Instant Policy Iteration}.

The experimental procedure was structured as follows:
\begin{enumerate}
    \item \textbf{10-Minute Warm-up:} Each participant undergoes a brief warm-up session to familiarize themselves with the device operation.
    \item \textbf{Offline PI:} Participants collect 5 trajectories using the Offline (Policy Iteration) PI workflow.
    \item \textbf{Instant PI:} Participants collect 5 trajectories using our proposed Robot-Free Instant Policy Iteration workflow.
    \item \textbf{Feedback:} Upon completion, participants fill out a questionnaire. The corrective data collected is also stored for subsequent visualization analysis.
\end{enumerate}
To eliminate the bias, the order in which each user performs Offline PI and Instant PI is randomized.

\subsection{Qualitative Results}

\begin{figure}[h]
    \centering
    \includegraphics[width=\linewidth]{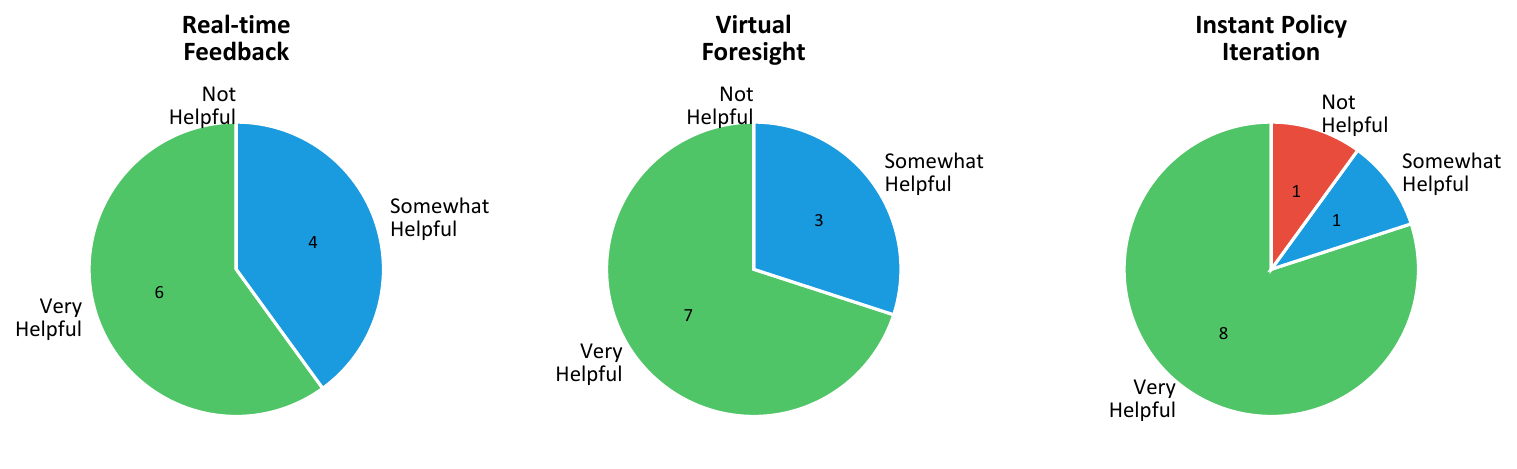}
    \caption{\textbf{User Study Results.} Visualization of user feedback regarding the helpfulness of Real-time Feedback, Virtual Foresight, and the Instant Policy Iteration workflow.}
    \label{fig:user_survey_results}
\end{figure}

The results of our user survey are summarized in \cref{fig:user_survey_results}.
Regarding the system's core features:
\begin{itemize}
    \item \textbf{Real-time Feedback:} All users report that the real-time constraints and feedback mechanisms are beneficial for data collection, with over half of the participants rating them as ``Very Helpful".
    \item \textbf{Virtual Foresight:} There is a consensus among all users that our Virtual Foresight feature effectively assists in discovering model failure cases. Notably, 7 out of 10 participants rate it as ``Very Helpful".
    \item \textbf{Instant Policy Iteration:} Crucially, 8 out of 10 users believe that the Instant Policy Iteration is highly beneficial for realizing the model's improvement.
\end{itemize}
When asked about the most significant advantage of RoboPocket, the majority of users provide feedback that the ``Real-time Uploading" and ``Online Finetuning" mechanisms make the collection process significantly more effective.

\subsection{Visualization Analysis}

\begin{figure}[h]
    \centering
    \includegraphics[width=0.5\linewidth]{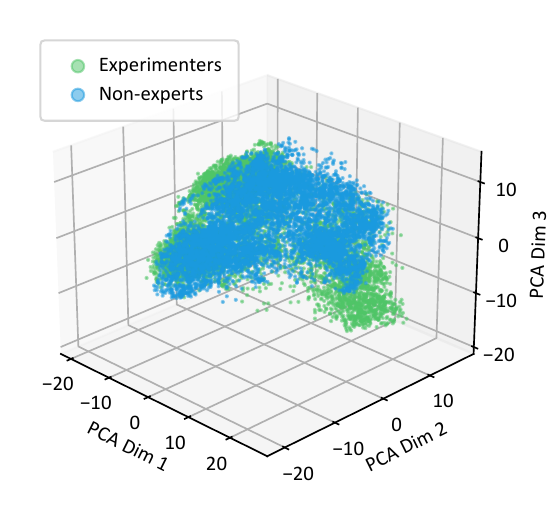}
    \caption{\textbf{State Coverage Visualization.} PCA projection of DINOv2 features comparing the coverage of data collected by non-experts using RoboPocket versus expert data.}
    \label{fig:state_coverage}
\end{figure}

To assess the quality of the data collected by non-experts, we perform a visualization analysis. We utilize PCA to project the DINOv2~\cite{oquab2023dinov2} features of the state data and compare non-expert data with the data collected by experimenters.
Here, we use the Offline PI data for comparison since each participant’s online iterations are limited.
As shown in \cref{fig:state_coverage}, our Robot-Free Policy Iteration enables non-expert users to achieve a state coverage comparable to that of experienced experimenters.

\begin{figure}[h]
    \centering
    \includegraphics[width=\linewidth]{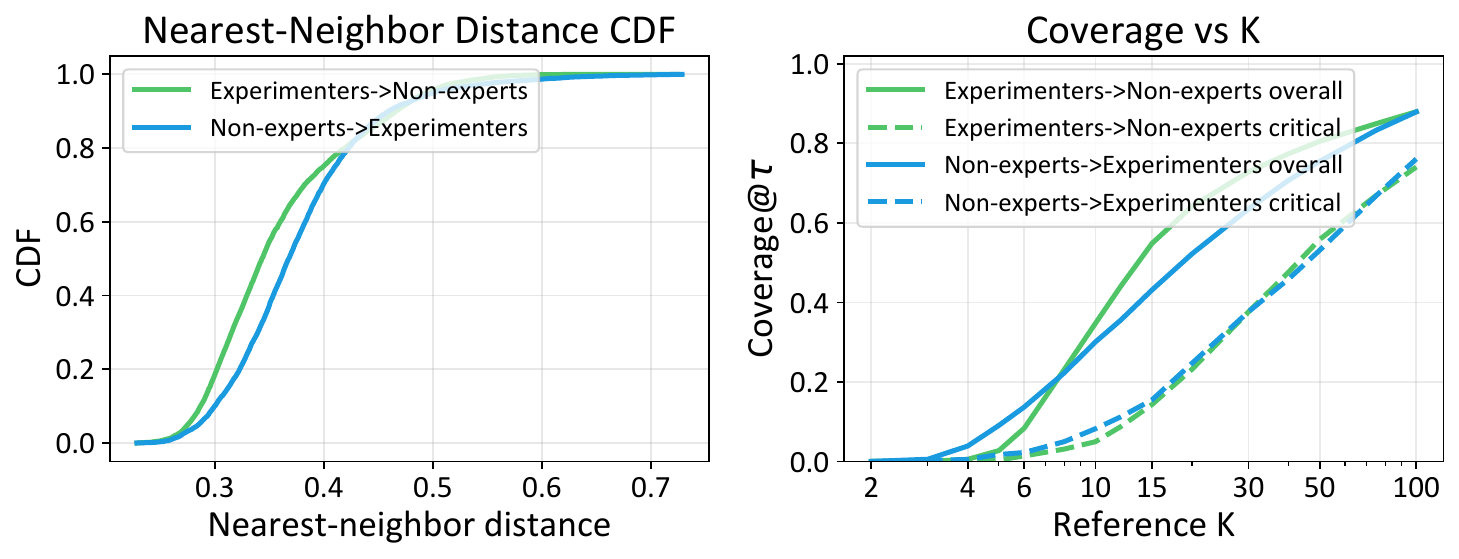}
    \caption{\textbf{Quantitative State Coverage Analysis.} Cross-dataset KNN coverage between non-expert and expert data in the PCA feature space.}
    \label{fig:state_coverage_knn}
\end{figure}

We further quantify state coverage in the 3D PCA feature space using cross-dataset $K$-nearest-neighbor (KNN) analysis. As shown in \cref{fig:state_coverage_knn}, most states from one dataset fall within the neighborhood of the other dataset when $K{=}1$, and the coverage curves converge rapidly at larger $K$. This pattern also holds for critical states, defined as the top 10\% states with the highest feature variation, indicating that non-expert users can capture state distributions close to those collected by experienced experimenters.

These analyses support comparable coverage in the chosen visual feature space, including critical states, but do not directly compare the end-to-end performance of policies trained on expert and non-expert corrections. The limited number of online corrections per participant also prevents this study from isolating the effect of collector expertise on policy improvement. A direct policy-performance comparison remains necessary to establish that stronger conclusion.

\section{Data Collection Details}

In \citet{hu2024data}, the authors conduct a comprehensive empirical study to investigate how the generalization performance of robot policies scales with the number of training environments, objects, and demonstrations. They discover that policy performance follows a power-law relationship with data diversity, concluding that increasing the diversity of environments and objects is significantly more critical for zero-shot generalization than merely increasing the number of demonstrations per scene.

\begin{figure}[h]
    \centering
    \includegraphics[width=\textwidth]{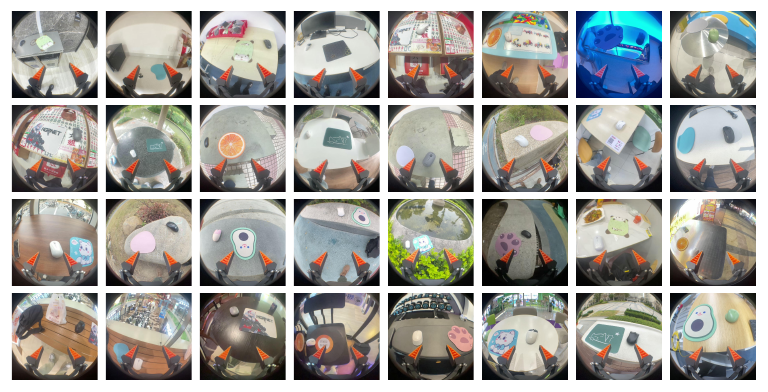}
    \caption{\textbf{Training environments for verifying Data Scaling Laws.} These environments encompass a wide range of lighting conditions and textures.}
    \label{fig:scalilng_enviroments}
\end{figure}

\begin{figure}[h]
    \centering
    \includegraphics[width=\textwidth]{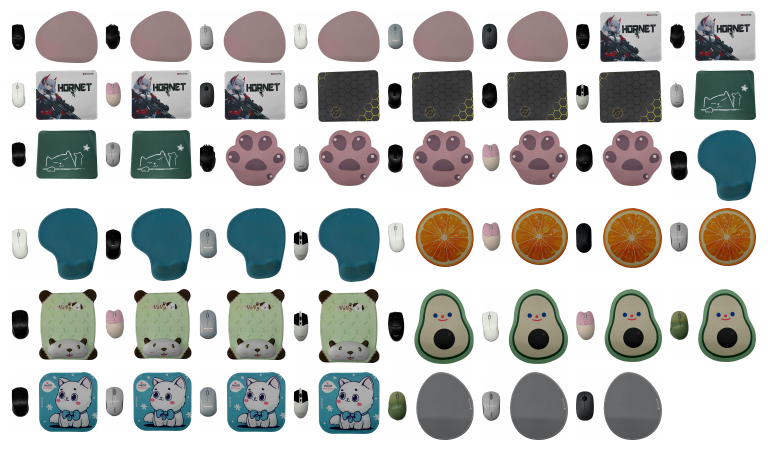}
    \caption{\textbf{Training object pairs for verifying the Data Scaling Laws.} We combine mice and mouse pads to obtain a diverse set of object pairs.}
    \label{fig:scaling_objects}
\end{figure}

Following their experimental protocol, we collect data for the ``Mouse Arrangement'' task (see \cref{subsec:system_verification} and \cref{fig:scaling_laws}) to verify that our RoboPocket system is capable of generating high-quality data that adheres to these established scaling laws.
We collect training data across 32 environments and 47 object pairs, which are shown in \cref{fig:scalilng_enviroments} and \cref{fig:scaling_objects}, respectively.
We randomly take 2 object pairs for data collection in 1 environment and collect 25 demonstrations for 1 env-object pair.
To broaden the coverage of the dataset, the environments include both indoor and outdoor settings with diverse lighting conditions and textures. The mice and mouse pads are combined to generate object pairs.
The evaluation is done in 3 different scenes with 2 initial robot poses and 3 initial object poses.

\section{Scoring Details}
\label{appendix:scoring}

For all experiments, we use normalized scores to evaluate policy performance. The specific scoring criteria for each task are detailed below:

\paragraph{Mouse Arrangement}
We adopt similar scoring criteria to Data Scaling Laws. The robot is required to pick up a mouse and then place it on a mouse pad.
The maximum score is 6 points, with 3 points for picking and 3 points for placing.

\noindent\textbf{Picking:}
\begin{itemize}
    \item \textbf{0 points:} The gripper fails to approach the mouse.
    \item \textbf{1 point:} The gripper contacts the mouse, but drops it immediately during the lift.
    \item \textbf{2 points:} The gripper displaces the mouse significantly before grasping, or successfully grasps the mouse but drops it at a higher height.
    \item \textbf{3 points:} The gripper successfully grasps the mouse.
\end{itemize}

\noindent\textbf{Placing:}
\begin{itemize}
    \item \textbf{0 points:} The gripper hovers without approaching the mouse pad, or releases the mouse prematurely from a height.
    \item \textbf{1 point:} The mouse lands outside the mouse pad or flips over due to an excessive release height.
    \item \textbf{2 points:} The mouse rests only partially on the pad, or exhibits bouncing or shifting caused by a high release point.
    \item \textbf{3 points:} The gripper descends and places the entire mouse securely and stably on the pad.
\end{itemize}

\paragraph{Block Sorting}
The robot is required to sort the Red (R), Green (G), and Blue (B) blocks into their corresponding boxes in a sequential order. The maximum score is 9 points, with 3 points allocated for each block. The detailed breakdown is:
\begin{itemize}
    \item \textbf{0 points:} The gripper moves to an area without blocks and closes.
    \item \textbf{1 point:} The gripper attempts to approach the correct block but fails to grasp it successfully.
    \item \textbf{2 points:} The gripper successfully grasps the correct block but places it into the wrong box.
    \item \textbf{3 points:} The gripper grasps the correct block and places it into the correct box.
\end{itemize}
If the model completes the sorting in an incorrect order, the score is calculated based on the Longest Increasing Subsequence (LIS) matching the ground truth. For example, completing the task in R-B-G order yields 6 points, while B-G-R yields 3 points.

\paragraph{Seasoning Pouring}
The task involves picking up seasoning jars in a left-to-right order, pouring them over a plate, and placing them back on the table. The maximum score is 12 points, with 4 points for each jar.
\begin{itemize}
    \item \textbf{0 points:} The gripper moves to an area without jars and closes.
    \item \textbf{1 point:} The gripper attempts to approach the correct jar but fails to grasp it.
    \item \textbf{2 points:} The gripper grasps the correct jar but fails to reach the pouring position above the plate or drops the jar.
    \item \textbf{3 points:} The gripper successfully pours the seasoning but fails to place the jar stably back on the table.
    \item \textbf{4 points:} The gripper successfully pours and places the jar stably on the table.
\end{itemize}
Similar to Block Sorting, if the execution order is incorrect, the final score is calculated based on the LIS match.

\paragraph{Towel Folding}
The robot must grasp the towel's corners and place them in the correct positions to complete two folds. The maximum score is 6 points, with 3 points per fold.
\begin{itemize}
    \item \textbf{0 points:} The gripper moves to a wrong target location.
    \item \textbf{1 point:} The gripper moves to the correct corner but fails to grasp the towel.
    \item \textbf{2 points:} The gripper grasps the correct corner but fails to place it in the correct position.
    \item \textbf{3 points:} The gripper grasps the correct corner and places it in the correct position.
\end{itemize}

\paragraph{Snack Bagging (Bimanual)}
The left arm holds the bag while the right arm sequentially places four snacks into it. The maximum score is 12 points, with 3 points for each snack.
\begin{itemize}
    \item \textbf{0 points:} The gripper moves to a wrong target location.
    \item \textbf{1 point:} The gripper moves to the correct snack but fails to grasp it.
    \item \textbf{2 points:} The gripper grasps the snack but fails to put it into the bag.
    \item \textbf{3 points:} The gripper grasps the snack and successfully puts it into the bag.
\end{itemize}

\section{Real Robot Experiment Hardware Setup}
\label{appendix:hardware}

\begin{figure}[h]
    \centering
        \includegraphics[width=0.63\linewidth]{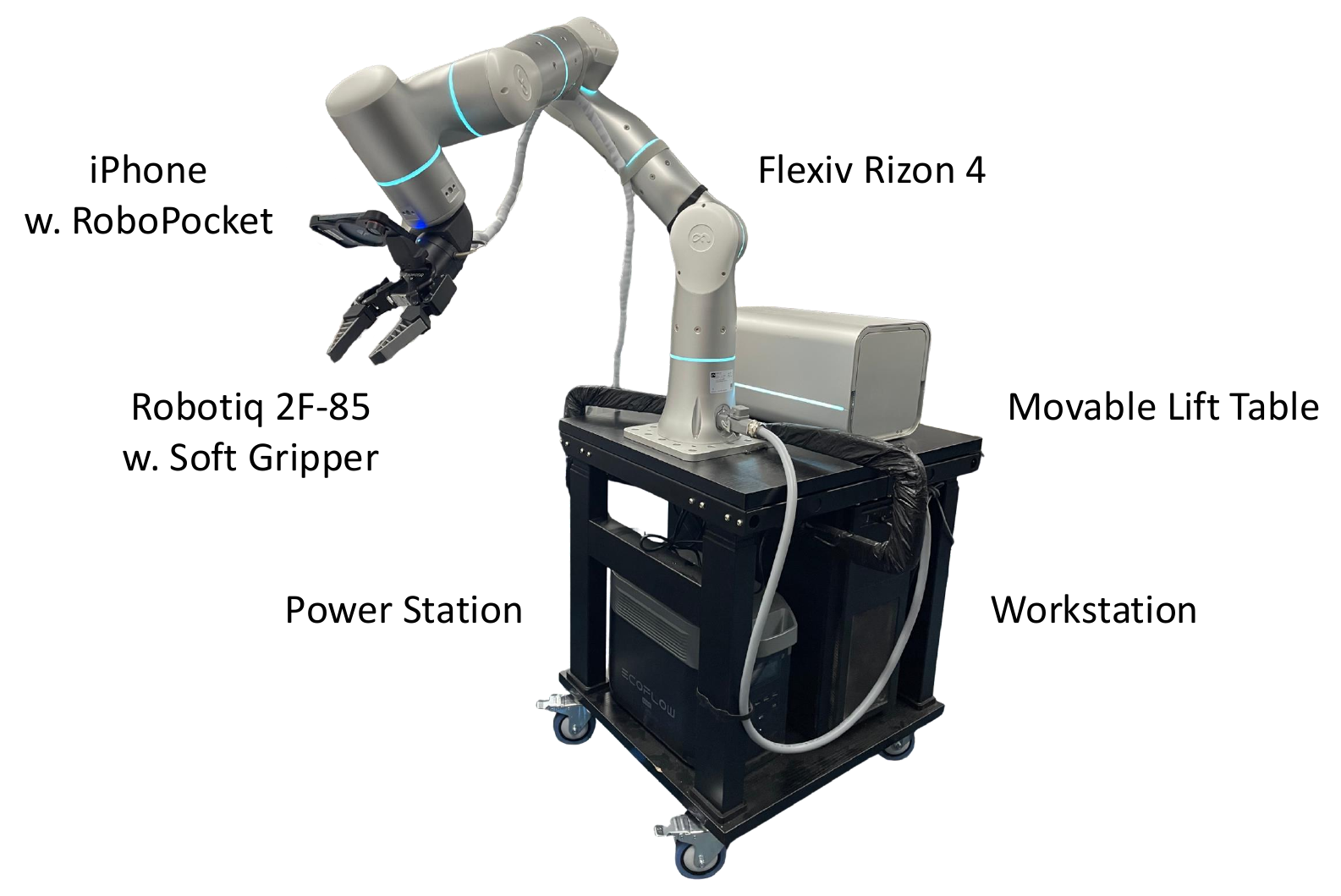}
    \caption{\textbf{Robot Evaluation Setup.} The setup includes a movable lifting table, a Flexiv Rizon 4 robot with a Robotiq 2F-85 gripper, and a workstation powered by a mobile power station.}
    \label{fig:hardware_setup}
\end{figure}

Following Data Scaling Laws~\cite{hu2024data}, we construct a custom movable lifting table.
As shown in \cref{fig:hardware_setup}, the setup features a Flexiv Rizon 4 robot arm~\cite{flexiv_rizon4} equipped with a Robotiq 2F-85 adaptive gripper~\cite{robotiq_2f85}, connected via a 90-degree adapter.
To ensure physical isomorphism with our handheld device and maintain adaptive grasp performance, the gripper is fitted with the same TPU soft fingers used in the RoboPocket collector.
Notably, in the Data Scaling Laws experiments (see \cref{subsec:system_verification} and \cref{fig:scaling_laws}), we use the DH-ROBOTICS AG-160-95 gripper and a corresponding isomorphic handheld collector.
Additionally, an iPhone mount is attached to the gripper, where the RoboPocket APP streams camera feeds to a workstation in real-time.
The workstation, powered by an Intel Core i9-12900K CPU and an NVIDIA GeForce RTX 3090 GPU, is housed on the table's lower shelf.
This workstation also acts as the Data Serving Node and the Training Server.
The entire system is powered by an EcoFlow DELTA 3 MAX portable power station.
For bimanual tasks, two such platforms can be arranged side-by-side to support dual-arm manipulation.

During \textit{Robot-free Instant Policy Iteration}, we use another workstation as an Inference Server, which is equipped with an Intel Core i9-13900K CPU and an NVIDIA GeForce RTX 4090 GPU.

\section{Policy Training Details}
\label{appendix:training}

We use the CNN-based Diffusion Policy based on Diffusion Policy~\cite{chi2025diffusion} and UMI~\cite{chi2024universal} code base for policy training.
We set the observation horizon to $T_{\text{obs}} = 1$ so that observations collected during \textit{Robot-free Instant Policy Iteration} exclude any explicit or implicit velocity information. This design prevents instability in human motion speed from being encoded into the observations.
Although we discard historical context, large-scale results~\cite{intelligence2025pi} have confirmed that it does not impose an excessive impact on the scalability.
All data is originally collected at 30Hz but is slowed down by a factor of 3 to 10Hz to align with the robot's physical joint velocity constraints.
The specific hyperparameters used for training are listed in \cref{tab:hyperparameters}.

\begin{table}[h]
    \centering
    \caption{Hyperparameters for Policy Training}
    \label{tab:hyperparameters}
    \begin{tabular}{ll}
        \toprule
        \textbf{Hyperparameter} & \textbf{Value} \\
        \midrule
        Observation Horizon ($T_{obs}$) & 1 \\
        Action Prediction Horizon ($T_{pred}$) & 16 \\
        Action Execution Horizon ($T_{exec}$) & 8 \\
        Training Epochs & 600 \\
        Batch Size & 64 \\
        Optimizer & AdamW \\
        Optimizer Momentum & $\beta_1=0.95, \beta_2=0.999$ \\
        U-Net Learning Rate & $3.0 \times 10^{-4}$ \\
        \multirow{2}{*}{Observation Encoder} & DINOv2 (Towel Folding)~\cite{oquab2023dinov2} \\
         & CLIP (Others)~\cite{radford2021learning} \\
        Observation Encoder Learning Rate & $3.0 \times 10^{-5}$ \\
        Learning Rate Schedule & Cosine Decay \\
        Training Denoising Steps & 50 \\
        Inference Denoising Steps & 16 \\
        \bottomrule
    \end{tabular}
\end{table}

During the \textit{Robot-Free Instant Policy Iteration}, we adjust several training parameters, which are summarized in \cref{tab:pi_hyperparameters}.

\begin{table}[h]
    \centering
    \caption{Hyperparameters for Robot-free Instant Policy Iteration}
    \label{tab:pi_hyperparameters}
    \begin{tabular}{ll}
        \toprule
        \textbf{Hyperparameter} & \textbf{Value} \\
        \midrule
        Batch Size & 32 \\
        Learning Rate & $1.0 \times 10^{-4}$ \\
        Observation Encoder Learning Rate & $1.0 \times 10^{-5}$ \\
        Learning Rate Schedule & Constant \\
        Model Sync Interval $N$ & 100 Steps \\
        \bottomrule
    \end{tabular}
\end{table}

\end{document}